\documentclass[letterpaper]{article} % DO NOT CHANGE THIS
\usepackage{aaai25}  % DO NOT CHANGE THIS
\usepackage{times}  % DO NOT CHANGE THIS
\usepackage{helvet}  % DO NOT CHANGE THIS
\usepackage{courier}  % DO NOT CHANGE THIS
\usepackage[hyphens]{url}  % DO NOT CHANGE THIS
\usepackage{graphicx} % DO NOT CHANGE THIS
\usepackage{natbib}  % DO NOT CHANGE THIS AND DO NOT ADD ANY OPTIONS TO IT
\usepackage{caption} % DO NOT CHANGE THIS AND DO NOT ADD ANY OPTIONS TO IT
\usepackage{amsmath} 
\usepackage{graphicx}
\usepackage{booktabs}
\usepackage{booktabs}
\usepackage{subcaption}
\usepackage{color}
\usepackage{algorithm}
\usepackage{algorithmic}
\usepackage{tabularx}
\usepackage{booktabs}
\usepackage{multirow}
\usepackage{adjustbox}
\usepackage{caption}
\usepackage{xcolor}
\usepackage{newfloat}
\usepackage{listings}
\DeclareCaptionStyle{ruled}{labelfont=normalfont,labelsep=colon,strut=off} % DO NOT CHANGE THIS
\floatstyle{ruled}
\newfloat{listing}{tb}{lst}{}
\floatname{listing}{Listing}
\title{DAQ: Density-Aware Post-Training Weight-Only Quantization For LLMs}
\author{
    Yingsong Luo, Ling Chen\thanks{Corresponding Author}
}
\affiliations{
	State Key Laboratory of Blockchain and Data Security, College of Computer Science and Technology\\ Zhejiang University, Hangzhou, China \\
	\{yingsongluo, lingchen\}@cs.zju.edu.cn
}

\usepackage{bibentry}
\begin{document}
	
\maketitle

\begin{abstract}
	
	Large language models (LLMs) excel in various tasks but face deployment challenges due to hardware constraints. We propose \underline{d}ensity-\underline{a}ware post-training weight-only \underline{q}uantization (DAQ), which has two stages:  1) density-centric alignment, which identifies the center of high-density weights and centers the dynamic range on this point to align high-density weight regions with floating-point high-precision regions; 2) learnable dynamic range adjustment, which adjusts the dynamic range by optimizing quantization parameters (i.e., scale and zero-point) based on the impact of weights on the model output. Experiments on LLaMA and LLaMA-2 show that DAQ consistently outperforms the best baseline method, reducing perplexity loss by an average of 22.8\% on LLaMA and 19.6\% on LLaMA-2. Our code is available at \url{https://github.com/LuoYingSong/DAQ}.
	
\end{abstract}


\section{Introduction}

In recent years, large language models (LLMs) based on transformers \cite{transformer} have demonstrated remarkable performance in various natural language processing benchmarks \cite{gpt4, llama, llama2, opt}. These models exhibit deep semantic understanding and reasoning capabilities by learning from massive amounts of text. They often have billions of parameters, e.g., LLaMA-2 \cite{llama2} has up to 70 billion parameters. The immense size of these models leads to extremely high memory capacity requirements. In addition, recent studies \cite{squeezellm, gptq, awq} identify memory bandwidth as a primary bottleneck for LLMs \textcolor{black}{small-batch} inference.

\begin{figure}[htbp]
	\centering
	\begin{subfigure}[b]{0.49\textwidth}
		\includegraphics[width=\textwidth]{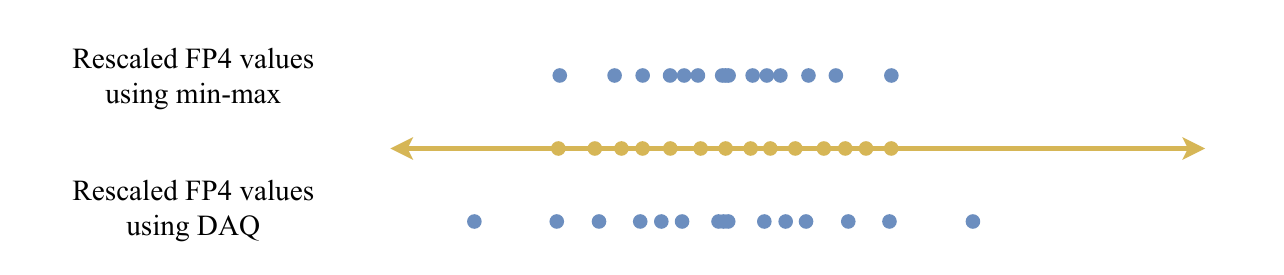}
		\caption{Low degree of outlierness}
		\label{fig::Asymmetric}
	\end{subfigure}
	\begin{subfigure}[b]{0.49\textwidth}
		\includegraphics[width=\textwidth]{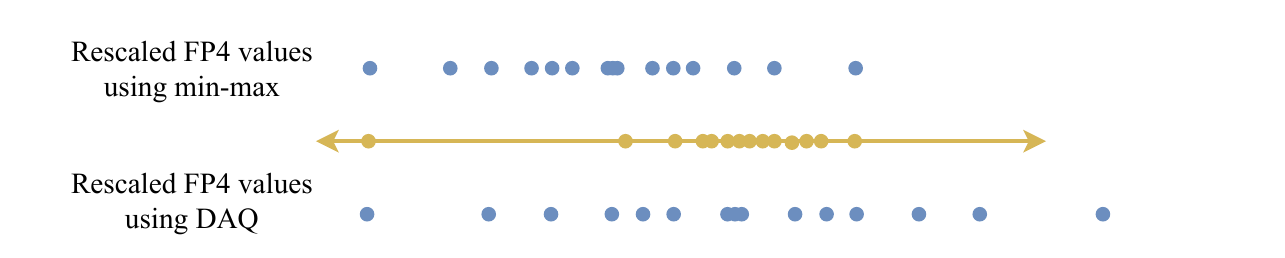}
		\caption{Asymmetric}
		\label{fig::low}
	\end{subfigure}
	\caption{Using two dynamic ranges \textcolor{black}{under} two weight distributions. The yellow points represent the original weights.  Under specific weight distributions, DAQ can \textcolor{black}{expand and shift} the dynamic range to align high-density weight regions with FP high-precision regions.}
	\label{fig:scale-affine-quantization}
\end{figure}

Although numerous model compression methods (e.g.,  quantization-aware training \cite{llm-qat}, pruning \cite{pruning}, and knowledge
distillation \cite{distillation}) can alleviate the memory demands, these methods require retraining for model compression. For LLMs with hundreds of billions of parameters, computational resources and the data requirements of retraining can be prohibitively expensive. In contrast, post-training quantization (PTQ) eliminates the need for model retraining, making it a promising solution in resource-constrained environments.

Mainstream PTQ methods \cite{gptq, awq} quantize weights or activations into uniform integers. The presence of outliers in transformer weights leads to a significant waste of intermediate quantization levels. Floating-point (FP) data types (e.g., FP4, NF4 \cite{qlora}, and AF4 \cite{af4}) have high precision at low numerical ranges and low precision at high numerical ranges. Therefore, FP quantization \cite{llm-fp4,mofq,fp8quantization, afpq} can better handle the outliers in transformer weights.

Determining the dynamic range is a crucial step in PTQ. 
Traditionally, the dynamic range is determined by the minimum and maximum values of weights or activations. As shown in Figure 1, considering the non-uniform properties of FP representation, under specific weight distributions, this method may hinder the mapping of high-density weight regions to FP high-precision regions. Recently, some studies \cite{towardacc, quanttune, adaquant} attempt to narrow the dynamic range to improve quantization precision by truncating outliers. However, some studies \cite{spqr, awq, owq} indicate that certain outliers have a more salient impact on model performance, and truncating these outliers may significantly impair the performance of the quantized model.

To address the aforementioned issues and fully leverage the non-uniform properties of FP representation, we propose \underline{d}ensity-\underline{a}ware post-training weight-only \underline{q}uantization (DAQ), which has two stages: 1) density-centric alignment (DCA), which aligns high-density weight regions with FP high-precision regions; 2) \textcolor{black}{learnable dynamic range adjustment (LDRA), which adjusts the dynamic range} based on the model output. In contrast to existing works that solely focus on narrowing the dynamic range to reduce quantization loss, we additionally consider expansion and \textcolor{black}{shift} the dynamic range as a potential optimization direction. The main contributions of this work are summarized as follows:

\begin{enumerate}
	\item We propose \textcolor{black}{DCA}, which \textcolor{black}{identifies the center of high-density weights and centers the dynamic range on this point to align high-density weight regions with FP high-precision regions.} 
	\item We propose \textcolor{black}{LDRA}, which adjusts the dynamic range by optimizing quantization parameters (i.e., scale and zero-point) based on the output of the original model and the quantized model.
	\item We evaluate our method on LLaMA and LLaMA-2, and the results demonstrate that DAQ outperforms the best PTQ method by an average of 22.8\% on LLaMA and 19.6\% on LLaMA-2 in terms of perplexity loss.
\end{enumerate}

\section{Related \textcolor{black}{Work}}
% According to the object of quantization, existing PTQ methods can be categorized into two types: weight-activation quantization and weight-only quantization.
\textcolor{black}{In this section, we review two main categories of relevant research in PTQ. Specifically, we first discuss weight-activation quantization, followed by an overview of weight-only quantization. This review situates our work within the existing methods and highlights the significance of our proposed DAQ.}

\subsection{Weight-\textcolor{black}{A}ctivation Quantization}
Weight-activation quantization methods attempt to quantize both weights and activations, aiming to utilize specialized INT8 general matrix multiplication kernels, which can reduce computational requirements by up to 50\% compared to FP16 kernels \cite{trt}. These works focus on addressing the challenge posed by the highly-dynamic intermediate activations of LLMs. The ranges of different channels often exceed a thousandfold, resulting in a significant performance drop. To address this challenge, researchers propose various methods. For example, SmoothQuant \cite{smoothquant} and Outlier Suppression \cite{os} employ channel-wise quantization to handle the large dynamic range of activations caused by a coarse quantization granularity; OmniQuant \cite{omni} and PaC \cite{towardacc} propose learnable boundaries of dynamic range to mitigate the large dynamic range of activations caused by outliers. \textcolor{black}{However, these methods typically can only quantize weights to 8 bits, and lower bit width leads to a significant model performance degradation, which is insufficient to effectively address the memory bottleneck caused by weight access in small-batch inference \cite{awq}.}

\subsection{Weight-Only Quantization}

To further alleviate the memory bottleneck of \textcolor{black}{small-batch inference}, numerous researchers attempt to quantize the weights only. Weight-only quantization speeds up inference by reducing the \textcolor{black}{size} of weight access. In this scenario, the primary challenge is to address the degradation in quantization precision caused by outliers when using 4 bits or lower bits for weight representation \cite{kbit}. For example, AdaQuant \cite{adaquant}, LAPQ \cite{lapq}, and  ACIQ \cite{aciq} focus on truncating outliers by optimizing quantization parameters; SpQR \cite{spqr}, OWQ \cite{owq}, and AWQ \cite{awq} emphasize that salient weights have more impact on the activations. To protect salient weights, SpQR and OWQ isolate salient weights and store them with FP16, while AWQ proposes a hardware-friendly method by utilizing a per-channel scaling method.

Due to the limited expressiveness of integer quantization, some researchers explore how to utilize FP quantization to improve the performance of previous methods. For example, MoFQ \cite{mofq} determines the optimal data type from various candidates (e.g., INT4 and FP4) for each layer based on tensor error;  AFPQ \cite{afpq} employs separate scales for positive and negative weights, addressing the asymmetric distributions commonly found in weights. However, these methods usually determine the dynamic range based on the minimum and maximum values or narrow said dynamic range to truncate outliers, which may not fully leverage the non-uniform properties of FP representation, particularly in cases where expanding or shifting the dynamic range could be beneficial. 

In this work, we propose a dynamic range optimization method for FP quantization, which considers adjusting the dynamic range to align high-density weight regions with FP high-precision regions and then adjusts the dynamic range based on the impact of weights on the model output.

\section{Preliminaries}
% \textcolor{black}{In this section, we introduces key concepts and notations for quantization and dequantization. Then, We discuss the optimization objective in weight-only quantization.}

% \textbf{}
\subsubsection{Quantization.}
Quantization aims to represent model weights or activations with lower bit-width representations. The basic idea is to map continuous real numbers to a finite set, and this mapping is typically implemented as a function. For given original weights $W$, the quantized weights \textcolor{black}{$W_\text{q}$} can be calculated as follows:
\begin{equation}
	\begin{aligned}
		\textcolor{black}{W_\text{q}} &= \operatorname{RTN}\left(\frac{W}{s} + z\right)  
	\end{aligned}
	\label{q_formal}
\end{equation}
where $\operatorname{RTN}$ represents the round-to-nearest function, $s$ is the scale, and $z$ is the zero-point. The quantization parameters $s$ and $z$ can be calculated as follows:
\begin{equation}
	\begin{aligned}
		s &= \frac{\beta - \alpha}{x_\text{max} - x_\text{min}}\\
		z &= x_\text{min} - \frac{\alpha}{s}
	\end{aligned}
	\label{s_formal}
\end{equation}
where $[\alpha, \beta]$ is the dynamic range, and $[x_\text{min}, x_\text{max}]$ is the quantization range, which represents the range of the quantized values.

\subsubsection{Dequantization.} Dequantization aims to restore the quantized weights to their original bit-width during the inference process. The restored weights $\tilde{W}$ can be calculated as follows:
\begin{equation}
	\begin{aligned}
		\tilde{W} &= \textcolor{black}{s(W_\text{q}} - z) 
	\end{aligned}
	\label{dq_formal}
\end{equation}

\subsubsection{Optimization Objective in Weight-Only Quantization.}
The optimization objective in weight-only quantization is to find  the optimal zero-point $z^*$ and scale $s^*$    that minimize the quantization loss $\mathcal{L}$, which can be formulated as follows:
\begin{equation}
	z^*, s^* = \underset{z,s}{\operatorname{argmin}  \mathcal{L}}
	\label{optimization}
\end{equation}

The quantization loss $\mathcal{L}$ is defined as the difference between the intermediate output of the quantized layer and the output of the original layer, which can be calculated as follows:
\begin{equation}
	\begin{aligned}
		\mathcal{L} = \left\| \tilde{W}X - WX \right\|_\text{F}^2
	\end{aligned}
	\label{loss}
\end{equation}
where  $X$ represents the input data of each layer obtained from the calibration dataset, and $\left\|.\right\|_\text{F}$ represents the Frobenius norm operator.

\section{Methodology}

% \subsection{Method Overview}
\textcolor{black}{In this section, we present our proposed DAQ, which} has two stages: 1) DCA \textcolor{black}{identifies the center of high-density weights and centers the dynamic range on this point to align high-density weight regions} with FP high-precision regions; 2) \textcolor{black}{LDRA} further adjusts the dynamic range by optimizing the quantization parameters based on the impact of weights on the model output.

\subsection{Density-Centric Alignment}

\begin{figure}[t]
	\centering
	\includegraphics[width=0.5\textwidth]{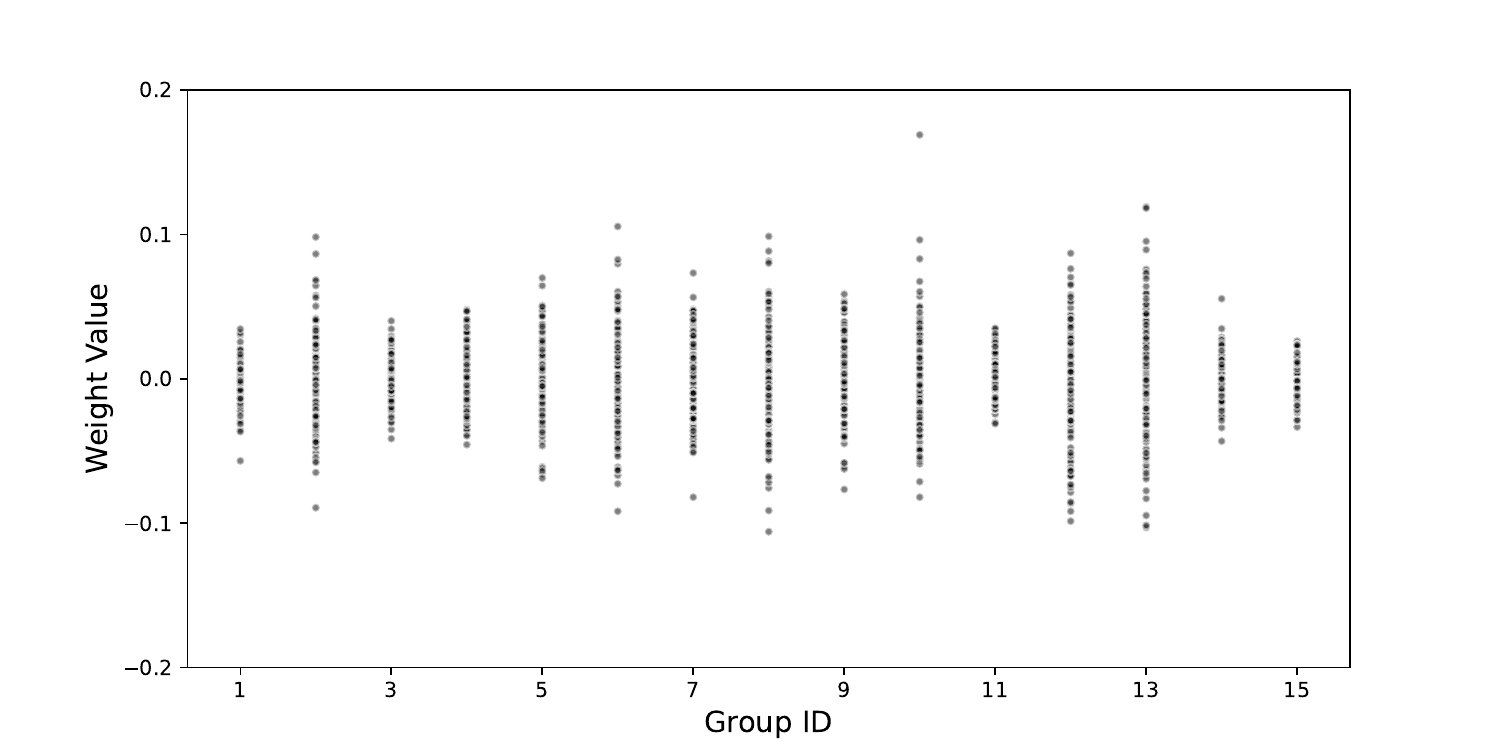} % 插入图片并设置宽度为文本宽度的50%
	\caption{The distribution of 15 randomly selected weight groups (group size: 128) from LLaMA-2-7B. \textcolor{black}{LLaMA-2-7B is a state-of-the-art LLM with 7 billion parameters, known for its strong performance across various natural language processing tasks.}} % 图片标题
	\label{fig:beeswarm} % 图片标签，用于引用
\end{figure}

As shown in Figure \ref{fig:beeswarm}, the middle part of the weight distribution often exhibits a relatively high-density characteristic, while the weights on both sides of most groups are sparse and asymmetric. Thus, directly using the maximum and minimum values to determine the dynamic range will cause high-density weight regions to be mapped to FP low-precision regions.

To address this issue, we propose \textcolor{black}{DCA}, which aims to align high-density weight regions with FP high-precision regions. To find the center of high-density weights, we introduce the concept of the $p$-th quantile. Given a weight group $\mathbf{w}$, the $p$-th quantile is the value below which $p\%$ of the weights fall. the center of high-density weights $p_\text{c}$ can be calculated as follows:
\begin{equation}
	p_\text{c} = \frac{\operatorname{Quantile}(\mathbf{w}, m) + \operatorname{Quantile}(\mathbf{w}, 100-m)}{2}
\end{equation}
where $\operatorname{Quantile}$ represents the function that calculates the quantile of weights, and $m$ represents the specific clipping rate, which is usually a small value.

\textcolor{black}{After determining the center of high-density weights $p_\text{c}$, we introduce a variable $k$ to represent the maximum distance between $p_\text{c}$ and the extreme values of weights.} Let $w_\text{max}$ and $w_\text{min}$ denote the maximum and minimum values of $\mathbf{w}$, respectively. \textcolor{black}{The variable $k$ can be calculated as follows}:
\begin{equation}
	k = \operatorname{max}(w_\text{max} - p_\text{c}, p_\text{c} - w_\text{min})
\end{equation}

Subsequently, \textcolor{black}{we establish the dynamic range $[\alpha, \beta]$ by setting it to $[p_\text{c} - k, p_\text{c} + k]$, which centers the dynamic range on $p_\text{c}$ to align high-density weight regions with the FP high-precision regions while avoiding clipping of outliers. The quantization parameters are then determined based on this new dynamic range  $[\alpha, \beta]$ and Equation \eqref{s_formal}.}

\subsection{Learnable Dynamic Range Adjustment}

\begin{figure}[t]
	\centering
	\begin{subfigure}[b]{.49\textwidth}
		\includegraphics[width=\textwidth]{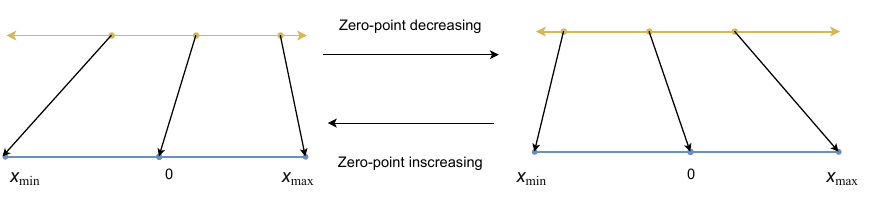}
		\caption{Adjusting the zero-point shifts the dynamic range left or right.}
		% \label{fig::Asymmetric}
	\end{subfigure}
	\hfill % 添加一些水平间距
	\begin{subfigure}[b]{.49\textwidth}
		\includegraphics[width=\textwidth]{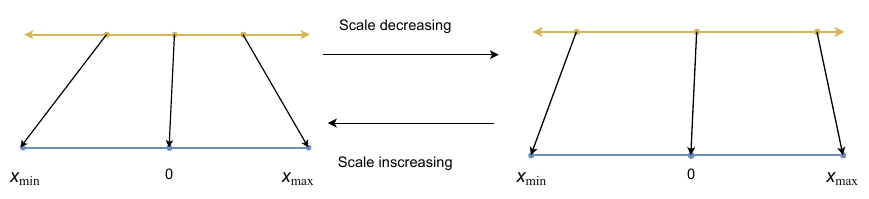}
		\caption{Adjusting the scale expands or contracts the dynamic range.}
		% \label{fig::low}
	\end{subfigure}
	\caption{The effect of adjusting zero-point and scale on the dynamic range. The yellow line represents the dynamic range, while the blue line represents the quantization range. }
	\label{fig:zp-scale-increasing}
\end{figure}

Although DCA can align high-density weight regions with the FP high-precision regions by adjusting the dynamic range, DCA still has two limitations: 1) The degree of outlierness varies among different weight groups. \textcolor{black}{As shown in Figure \ref{fig::Asymmetric}}, for weight groups with a low degree of outlierness, expanding the dynamic range can lead to higher precision; 2) Absence of considering the impact of weights on the model output may result in a suboptimal dynamic range. Therefore, it is necessary to further adjust the dynamic range.

Thus, we propose \textcolor{black}{LDRA}, a quantization parameter optimization method based on the finite difference method. Specifically, this method aims to adjust the dynamic range by optimizing the quantization parameters based on the impact of weights on the model output.

Optimizing quantization parameters to reduce quantization error has been used in previous works \cite{adaquant, lapq, aciq}. However, these methods are primarily designed for integer quantization, where expanding or shifting the dynamic range typically leads to decreased quantization precision. Consequently, these methods commonly employ saturation quantization, which involves decreasing the scale to narrow the dynamic range and filter outliers. In contrast, our method is tailored for FP quantization and based on the fact that the dynamic range is affected by quantization parameters, as shown in Figure 3. We aim to adjust the dynamic range \textcolor{black}{by} optimizing quantization parameters to achieve higher precision, \textcolor{black}{without restricting both expansion and shift} of the dynamic range, which can leverage the non-uniform properties of FP representation.

We reformulate the loss function in Equation \eqref{loss} to explicitly include the quantization parameters:
\begin{equation}
	\mathcal{L} = \left\| s (\operatorname{RTN}(\frac{W}{s} + z) - z )X - WX \right\|_\text{F}^2
\end{equation}
This formulation shows that by optimizing the quantization parameters $s$ and $z$, we can adjust the dynamic range, which in turn affects the quantization loss. Our goal is to find the optimal values of $s$ and $z$ that minimize this loss, effectively determining the best dynamic range for quantization.

To overcome the non-differentiability of the $\operatorname{RTN}$ function used in the quantization, we employ the finite difference method to estimate the gradient, which can be calculated as follows:
\begin{equation}
	\begin{aligned}
		\Delta \mathcal{L}(x, \varepsilon) &= \mathcal{L}(x + \varepsilon) - \mathcal{L}(x - \varepsilon)\\
		\nabla_x \mathcal{L} &\approx \Delta \mathcal{L}(x, \varepsilon) 
	\end{aligned}
\end{equation}
where $x$ represents the value to be optimized, namely, the quantization parameters, and $\varepsilon$ represents the step size of the finite difference method.

Due to the non-smoothness of the objective function caused by $\operatorname{RTN}$ function, there may be drastic changes in some numerical values. Consequently, choosing an appropriate learning rate $\eta$ becomes particularly complex, which in turn leads to difficulties in the convergence of gradient descent.

In recent years, sign-based gradient descent (SignGD) methods \cite{signgd} show good robustness, as they only focus on the direction of the gradient rather than the numerical value. To a certain extent, they can resist the impact of drastic changes in the numerical value of the non-smooth objective function in certain regions. The parameters of the $k$-th iteration $x_{k}$ can be optimized as follows:
\begin{equation}
	x_{k} = x_{k-1} - \eta \cdot \operatorname{sign}(\nabla_{x} \mathcal{L})
\end{equation}
where $\eta$ is the learning rate, a small positive number representing the step size of each update. \textcolor{black}{The $\operatorname{sign}$ function can be calculated as follows:
	\begin{equation}
		\operatorname{sign}(x) = \begin{cases}
			-1 & \text{if } x < 0 \\
			0 & \text{if } x = 0 \\
			1 & \text{if } x > 0
		\end{cases}
	\end{equation}
	This function extracts only the sign of the gradient, effectively normalizing the update step and making it independent of the gradient value.}

By using the direction of the gradient rather than its numerical value as the basis for the update step, SignGD effectively avoids the problem of unstable update step size caused by drastic fluctuations in the gradient \textcolor{black}{values}. In addition, SignGD simplifies the gradient calculation and reduces the computational complexity, thereby improving the efficiency of the overall optimization process.

The learning rate $\eta$ in SignGD should be chosen to balance the speed and stability of the optimization process. To further improve the efficiency and stability of the optimization, we introduce a learning rate decay mechanism within SignGD. This mechanism gradually reduces the learning rate as the number of iterations increases, allowing for rapid progress in the early stages of optimization and more refined updates when approaching the optimal solution. The learning rate $\eta_t$  of the $t$-th iteration can be calculated as follows:
\begin{equation}
	\eta_t = \frac{\eta_0}{1 + d \cdot t}
\end{equation}
where $\eta_0$ is the initial learning rate, $d$ is the decay coefficient, and $t$ is the number of iterations. The learning rate decay method can not only accelerate the convergence speed in the early stage but also avoid oscillation and non-convergence in the later stage of optimization.

\section{Experiments}
\textcolor{black}{In this section, we present comprehensive experiments to evaluate the effectiveness of DAQ. We first introduce the experimental settings, including models, datasets, and implementation details. Next, we compare DAQ with state-of-the-art methods, conduct ablation studies, examine performance under limited calibration data, and demonstrate the integration of DAQ with existing quantization methods.}

\subsection{Settings}

\subsubsection{Quantization.} Our study primarily focuses on weight-only quantization, as it can demonstrate the capability to largely maintain the performance integrity of LLMs \cite{gptq}. The integer (INT) and  NormalFloat (NF) \cite{qlora} are employed in the experiments. We utilize group-wise quantization, a method that allows for the independent optimization of quantization parameters for each weight group. This method is extensively adopted in the field of LLM quantization \cite{awq, gptq, afpq}.  The size of the weight group is set to 256 in the experiments except otherwise specified. We sample the calibration dataset from the Pile \cite{pile}. This dataset is employed to mitigate the risk of overfitting to any particular downstream domain, thereby enhancing the generalizability of quantization methods.

\subsubsection{Hyperparameters.} 
LDRA is initialized with a learning rate of 1e-3, coupled with a decay rate of 0.05, a configuration chosen to enhance training stability. The step size for the finite difference method is established at 1e-4, striking a balance between the accuracy of gradient estimation and numerical stability. In DCA, we set the parameter $p$ to 2.275, corresponding to 2$\sigma$ (two standard deviations in a normal distribution). This selection is made to effectively capture the center of high-density weights.

\subsubsection{Models.} Our methodology is rigorously evaluated using LLaMA \cite{llama} and LLaMA-2 \cite{llama2}. These models are selected because they are state-of-the-art open-source LLMs. The choice of LLaMA and LLaMA-2 allows for a comprehensive assessment of our quantization method across varying model scales.

\subsubsection{Evaluation.} Consistent with established literature, we primarily assess the quantized models using language modeling tasks, specifically perplexity evaluation on WikiText-2 \cite{wikitext2}. Perplexity is defined as the exponential of the cross-entropy loss:
\begin{equation}
	Perplexity = \exp\left(-\frac{1}{N}\sum_{i=1}^N \log p(x_i|x_{<i})\right)
\end{equation}
where $N$ is the number of tokens, and $p(x_i|x_{<i})$ is the predicted probability for the $i$-th token given the preceding tokens. This metric is chosen for its demonstrated reliability in reflecting LLM performance. To quantify the improvement of DAQ over baseline methods, we calculate the percentage reduction as follows:
\begin{equation}
	Improvement = \frac{loss_{\text{baseline}} - loss_{\text{DAQ}}}{loss_{\text{baseline}}} \times 100\%
\end{equation}
\textcolor{black}{where} the $loss$ for each method is the difference between its perplexity and the FP16 perplexity.

% \noindent\textbf{Evaluation} Consistent with established literature, we primarily assess the quantized models using language modeling tasks, specifically perplexity evaluation on WikiText-2 \cite{wikitext2}. This metric is chosen for its demonstrated reliability in reflecting LLM performance.  To quantify the improvement of DAQ over baseline methods, we calculate the percentage reduction as follows:
% \begin{equation}
	% Improvement = \frac{loss_{\text{baseline}} - loss_{\text{DAQ}}}{loss_{\text{baseline}}} \times 100\%
	% \end{equation}
% where the $loss$ for each method is the difference between its perplexity and the FP16 perplexity.

\subsubsection{Environment.} All experiments are conducted on a high-performance Linux cloud server equipped with an NVIDIA Tesla A100 80GB GPU, an Intel(R) Xeon(R) Platinum 8352V CPU, and 360GB of RAM. DAQ and MoFQ are implemented with PyTorch 2.1.0. The other baseline methods are implemented in \textcolor{black}{Intel Neural Compressor} framework.

\subsection{Comparison with State-of-the-Art Methods}

\begin{table}[t]
	\centering
	\small
	\setlength{\tabcolsep}{3pt}
	\begin{tabular*}{\columnwidth}{@{\extracolsep{\fill}}llccccc@{}}
		\toprule
		& & \multicolumn{2}{c}{LLaMA-2} & \multicolumn{3}{c}{LLaMA} \\ 
		\cmidrule(r){3-4} \cmidrule(l){5-7}
		& & 7B & 13B & 7B & 13B & 30B \\ 
		\midrule
		FP16& - & 5.46 & 4.89 & 5.67 & 5.09 & 4.09   \\
		\midrule
		\multirow{3}{*}{INT4} 
		&Vanilla RTN & 5.77 & 5.07 & 5.99 & 5.38 & 4.36  \\
		&GPTQ (ICLR 23) & 5.73 & 5.05 & 5.94 & 5.34 & 4.35   \\
		&AWQ (MLSys 24) & 5.68 & 5.03 & 5.88 & 5.29 & 4.32  \\
		\midrule
		\multirow{5}{*}{NF4}
		&Vanilla RTN & 5.69 & 5.02 & 6.02 & 5.31 & 4.31 \\
		&GPTQ (ICLR 23) & 5.66 & 5.01 & 5.89 & 5.26 & 4.28   \\
		&AWQ (MLSys 24) & \underline{5.65} & \underline{5.00} & \underline{5.86} & \underline{5.22} & \underline{4.25} \\
		&MoFQ (\textcolor{black}{a}rXiv 23) & 5.68 & 5.02 & 5.94 & 5.31 & 4.32  \\
		&DAQ & \textbf{5.60} & \textbf{4.97} & \textbf{5.81} & \textbf{5.20} & \textbf{4.23} \\
		\midrule
		\multicolumn{2}{l}{Improvement} & 26.3\% & 27.3\% & 26.3\% & 15.4\% & 12.5\% \\
		\bottomrule
	\end{tabular*}
	\caption{Comparison of 4-bit weight-only quantization methods: WikiText-2 perplexity across LLaMA and LLaMA-2. Lower perplexity indicates better performance. The best results are \textbf{bolded}. The second-best results are \underline{underlined}.}
	\label{compare}
\end{table}

\begin{table}[t]
	\centering
	\small
	\setlength{\tabcolsep}{3pt}
	\begin{tabular*}{\columnwidth}{@{\extracolsep{\fill}}llccccc@{}}
		\toprule
		& & \multicolumn{2}{c}{LLaMA-2} & \multicolumn{3}{c}{LLaMA} \\ 
		\cmidrule(r){3-4} \cmidrule(l){5-7}
		& & 7B & 13B & 7B & 13B & 30B \\ 
		\midrule
		FP16& - & 5.46 & 4.89 & 5.67 & 5.12 & 4.13 \\
		\midrule
		\multirow{3}{*}{INT3} 
		&Vanilla RTN & 7.13 & 5.69 & 7.47 & 7.11 & 5.02 \\
		&GPTQ (ICLR 23) & 6.72 & 5.42 & 7.13 & 6.48 & 4.95 \\
		&AWQ (MLSys 24) & 6.62 & 5.09 & 7.07 & 6.39 & 4.87 \\
		\midrule
		\multirow{5}{*}{NF3}
		&Vanilla RTN & 6.63 & 5.58 & 6.61 & 6.49 & 4.84 \\
		&GPTQ (ICLR 23) & 6.52 & 5.44 & 6.39 & 6.33 & 4.73 \\
		&AWQ (MLSys 24) & \underline{6.46} & \underline{5.35} & \underline{6.37} & \underline{6.29} & \underline{4.71} \\
		&MoFQ (arXiv 23) & 6.60 & 5.55 & 6.57 & 6.46 & 4.81 \\
		&DAQ & \textbf{6.39} & \textbf{5.30} & \textbf{6.33} & \textbf{6.28} & \textbf{4.69} \\
		\midrule
		\multicolumn{2}{l}{Improvement} & 7.00\% & 10.9\% & 5.71\% & 0.85\% & 3.45\% \\
		\bottomrule
	\end{tabular*}
	\caption{Comparison of 3-bit weight-only quantization methods: WikiText-2 perplexity across LLaMA and LLaMA-2. Lower perplexity indicates better performance. The best results are \textbf{bolded}. The second-best results are \underline{underlined}.}
	\label{compare2}
\end{table}

To validate the effectiveness of DAQ, we compare it with state-of-the-art quantization methods for LLMs:

\noindent\textbf{Vanilla RTN}: A basic quantization method that determines quantization parameters based on the original weight range and uses round-to-nearest for value mapping. It is widely used in various quantization scenarios due to its simplicity and minimal inference overhead.

\noindent\textbf{GPTQ} \cite{gptq}: A weight quantization method that uses approximate second-order information to reconstruct quantized weights.

\noindent\textbf{AWQ} \cite{awq}: An activation-aware quantization method that identifies and protects salient weight based on activation distributions. In addition, it uses grid search to determine optimal scaling values.

\noindent\textbf{MoFQ} \cite{mofq}: A mixed integer and FP quantization method that selects the best quantization data types per layer based on quantization loss.

To ensure fairness, all quantization \textcolor{black}{methods} use the same sampled calibration dataset.  For vanilla RTN, which does not require calibration data, we use grid search on the calibration set to find optimal quantization parameters. Baseline hyperparameters are set according to their \textcolor{black}{original} papers.

\textcolor{black}{
	Table \ref{compare} and Table \ref{compare2} show the performance of various quantization methods across different model sizes and data types. The following phenomena can be observed:
}
\begin{enumerate}

	\item NF data types generally yield better perplexity scores compared to their integer counterparts. This suggests that NF is more effective in preserving model performance for LLMs. 
	
	\item  The quantized larger models outperform full-precision smaller models. This suggests that quantizing a larger model can yield better performance than using a full-precision smaller model, potentially offering a more efficient way to deploy advanced language models in resource-constrained environments.
	
	\item  DAQ consistently outperforms other quantization methods, including vanilla RTN, GPTQ, AWQ, and MoFQ across both LLaMA and LLaMA-2. As model size increases from 7B to 30B parameters, the relative performance gains of DAQ remain consistent, indicating good scalability. This superior performance is attributed to utilizing the non-uniform properties of FP representation to effectively protect salient weights. Through its two process stages, DAQ ensures that most of the salient weights are mapped to the FP high-precision regions, thus better preserving the overall performance.
	
\end{enumerate}

% \begin{table}[htbp]
	% \centering
	% \caption{Comparison of weight-only quantization methods on LLaMA-2-7B with varying group sizes. g-1 represents channel-wise quantization, while g256, g128, and g64 indicate group sizes of 256, 128, and 64, respectively. The best baseline is underlined.}
	% \begin{tabular}{lllllll}
		% \toprule
		%      & & g-1 & g256 & g128 & g64 \\
		%  \\ 
		%      \midrule
		% FP16& - &  \multicolumn{4}{c}{5.46}  \\\
		% \midrule
		% \multirow{3}{*}{INT4} 
		% &RTN & 6.16 &5.77& 5.69& 5.65   \\\
		% &GPTQ &5.87 &5.73& 5.63 & 5.61   \\\
		% &AWQ & 5.81 &5.68& 5.62&  5.59 \\\
		% \midrule
		% \multirow{5}{*}{NF4}
		% &RTN & 5.93 &5.69& 5.73& 5.62 \\\
		% &GPTQ & 5.81 &5.66& 5.61 & 5.59 \\\
		% &AWQ& \underline{5.75} & \underline{5.65}& \underline{5.60}& \underline{5.58}\\\
		% &MoFQ  & 5.81 &5.68& 5.70 & 5.63  \\\
		% &DAQ & \textbf{5.64} & \textbf{5.60} & \textbf{5.57} &\textbf{5.54} \\
		% \midrule
		% \multicolumn{2}{l}{Improvement} & 26\% & 40\% & 43\% & 46\% \
		% \bottomrule
		% \end{tabular}
	% \label{compare3}
	% \end{table}

\begin{table}[t]
	\small
	\centering
	\setlength{\tabcolsep}{4pt}
	\begin{tabular}{lllllll}
		\toprule
		& & g-1 & g256 & g128 & g64 \\
		\midrule
		FP16& - & \multicolumn{4}{c}{5.46} \\
		\midrule
		\multirow{3}{*}{INT4}
		&Vanilla RTN & 6.16 &5.77& 5.69& 5.65 \\
		&GPTQ (ICLR 23)  &5.87 &5.73& 5.63 & 5.61 \\
		&AWQ (MLSys 24)  & 5.81 &5.68& 5.62& 5.59 \\
		\midrule
		\multirow{5}{*}{NF4}
		&RTN & 5.93 &5.69& 5.73& 5.62 \\
		&GPTQ (ICLR 23) & 5.81 &5.66& 5.61 & 5.59 \\
		&AWQ (MLSys 24)& \underline{5.75} & \underline{5.65}& \underline{5.60}& \underline{5.58}\\
		&MoFQ (arXiv 23) & 5.81 &5.68& 5.70 & 5.63 \\
		&DAQ & \textbf{5.64} & \textbf{5.60} & \textbf{5.57} &\textbf{5.54} \\
		\midrule
		\multicolumn{2}{l}{Improvement} &37.9\% & 26.3\% & 21.4\% & 33.3\% \\
		\bottomrule
	\end{tabular}
	\caption{Comparison of weight-only quantization methods on LLaMA-2-7B with varying group sizes. g-1 represents channel-wise quantization, while g256, g128, and g64 indicate group sizes of 256, 128, and 64, respectively.  Lower perplexity indicates a better model. The best results are \textbf{bolded}. The second-best results are \underline{underlined}.}
	\label{compare3}
\end{table}

To further demonstrate the effectiveness of our method, we conducted an additional experiment evaluating DAQ and other quantization methods on LLaMA-2-7B with different group sizes. This evaluation provides insights into how quantization granularity affects model performance. We present the performance of 4-bit quantization for various group sizes in Table \ref{compare3}, where g-1 represents channel-wise quantization, while g256, g128, and g64 indicate group sizes of 256, 128, and 64, respectively. The following tendencies can be discerned:

\begin{enumerate}
	\item DAQ consistently outperforms all other quantization methods across all group sizes. This demonstrates the robustness of DAQ in terms of different quantization granularities.
	
	\item As group size increases (from g64 to g-1), the weight dynamic range expands, increasing the presence of outliers. DAQ shows superior performance in these scenarios, especially in channel-wise quantization (g-1). This suggests the effectiveness of DAQ in optimizing larger dynamic ranges, a common challenge in a coarse quantization granularity.
\end{enumerate}

% The results in Table \ref{compare} lead to the following conclusions:

% RTN-FP significantly outperforms RTN-INT for LLMs, indicating that floating-point quantization better adapts to LLM weight distributions, resulting in lower quantization loss.
% Methods that only use the original weight range to determine quantization parameters (RTN-INT, RTN-FP) perform worse than other methods using the same data type. This suggests that considering only the weight range while ignoring distribution and output impact leads to greater quantization loss.
% DAQ outperforms other quantization methods, demonstrating the effectiveness of optimizing quantization parameters using weight distribution and calibration data.
\subsection{Ablation Study}

To validate the effectiveness of DCA and LDRA in DAQ, we compare DAQ with the following variants on LLaMA-2-7B using NF4:

\noindent\textbf{DAQ-MinMax}: This method removes both DCA and LDRA, using the maximum and minimum values of each weight group to determine quantization parameters without calibration data.

\noindent\textbf{DAQ-Percentil}: This method is similar to DAQ-MinMax, which uses percentile-based clipping \cite{trt} to narrow the dynamic range by truncating outliers in each weight group.

\noindent\textbf{DAQ-DCA}: \textcolor{black}{This method removes LDRA, using DCA to determine quantization parameters without further optimization.}

\noindent\textbf{DAQ-LDRA}: This method removes DCA, applying LDRA directly to quantization parameters obtained from DAQ-MinMax.

\noindent\textbf{DAQ w/o Zero-point}: This method is similar to DAQ, but only optimizes the scale parameter in LDRA, keeping the zero-point from DCA.

\noindent\textbf{DAQ w/o Scale}: This method is similar to DAQ, but only optimizes the zero-point parameter in LDRA, keeping the scale from DCA.

\begin{table}[t]
	\centering
	\small
	% \begin{adjustbox}{width=.45\textwidth}
		% \begin{tiny}
			\begin{tabular}{ll}
				\toprule
				Method & Perplexity(↓) \\
				\midrule
				DAQ-MinMax & 6.65 \\
				DAQ-Percentile & 6.89 \\
				DAQ-DCA & 6.37  \\
				DAQ-LDRA & \underline{5.63}  \\
				DAQ w/o Zero-point & 5.65 \\
				DAQ w/o Scale & 5.68 \\
				DAQ & \textbf{5.60} \\
				\bottomrule
			\end{tabular}
			% \end{tiny}
		\caption{Comparison of DAQ with different variant methods. Lower perplexity indicates better performance. The best results are \textbf{bolded}. The second-best results are \underline{underlined}. }
		\label{ablation}
		% \end{adjustbox}
\end{table}

\textcolor{black}{Table \ref{ablation} shows the performance of variants on LLaMA-2-7B. The following phenomena can be
	observed:}

\begin{enumerate}
	\item DAQ outperforms DAQ-LDRA, and DAQ-DCA outperforms DAQ-MinMax and DAQ-Percentile, demonstrating the effectiveness of DCA. This is mainly because DCA aligns high-density weight regions with FP high-precision regions, which is particularly beneficial for LLM weights that often follow a concentrated distribution with long tails. \textcolor{black}{In addition, by} focusing on the distribution rather than its extremes, DCA preserves more information in the quantization process, especially for the high-density weights.
	
	\item DAQ outperforms DAQ-DCA, and DAQ-LDRA outperforms DAQ-MinMax and DAQ-Percentile, justifying the effectiveness of LDRA. LDRA further adjusts the dynamic range by optimizing quantization parameters \textcolor{black}{based on the output of the original model and the quantized model.}
	
	\item DAQ-Percentile underperforms DAQ-MinMax, indicating that simply truncating outliers can be detrimental to model performance. This suggests that some outlier weights have a salient impact on the model output, aligning with recent findings in LLM quantization research. 
	
	\item DAQ outperforms both DAQ w/o Zero-point and DAQ w/o Scale, indicating that optimizing both scale and zero-point, i.e., considering \textcolor{black}{both expansion/contraction and shift} of dynamic range, is necessary to achieve the best model performance.
\end{enumerate}

\subsection{Limited Calibration Dataset Experiment}

To assess the effectiveness of DAQ under the limited calibration dataset, \textcolor{black}{we compare it to AWQ, the best PTQ method in our previous experiments}, across varying calibration dataset sizes.

\begin{figure}[t]
	\centering
	\includegraphics[width=0.48\textwidth]{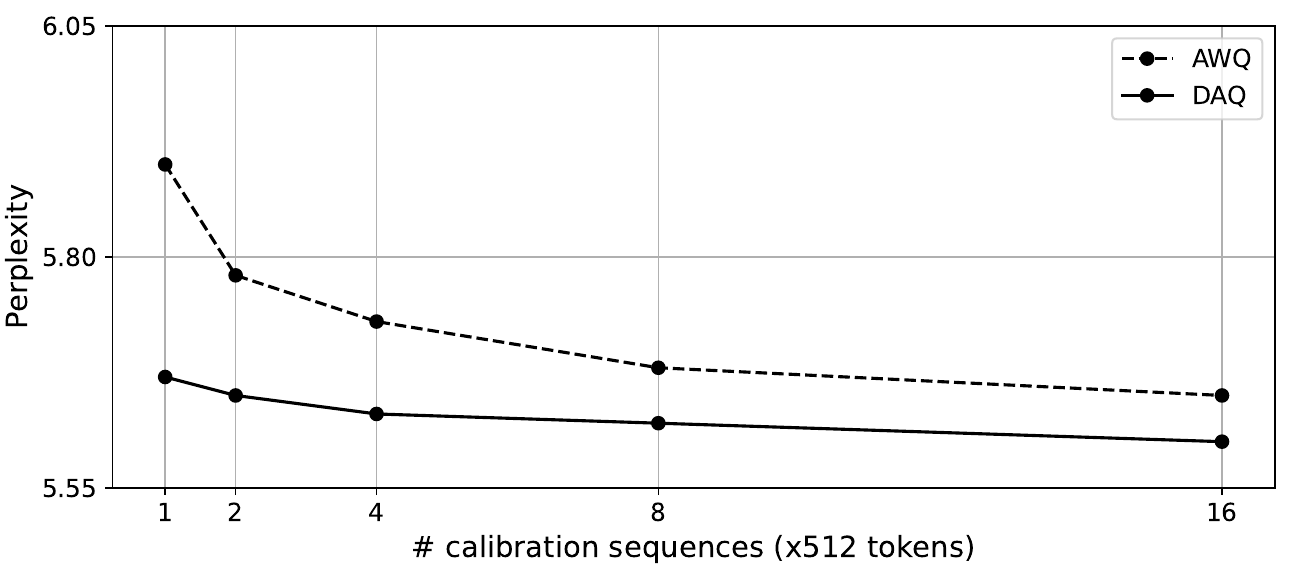}
	\caption{Perplexity under limited calibration datasets.}
	\label{fig:limited-dataset}
\end{figure}

As shown in Figure \ref{fig:limited-dataset}, DAQ consistently outperforms AWQ across all calibration dataset sizes. The performance gap is most significant with extremely limited data (1×512 tokens). Notably, DAQ using just 2×512 tokens achieves comparable performance to AWQ using 16×512 tokens, demonstrating its superior efficiency in utilizing limited calibration data. This superior performance is attributed to LDRA, which successfully determines an optimal dynamic range even with limited calibration data. 

\subsection{Integration with Existing Methods}

To demonstrate the versatility and potential synergies of DAQ, we conducted experiments combining our method with AWQ, a state-of-the-art weight-only quantization method.  DAQ employs the same inference process as vanilla RTN, focusing solely on optimizing quantization parameters. This design choice enables DAQ to seamlessly integrate with and enhance the performance of advanced quantization methods. We compared the performance of AWQ alone, DAQ alone, and the combined AWQ+DAQ using 4-bit NF quantization across various models on LLaMA-2.

% Table \ref{tab:integration_results} presents the results of this comparison:

\begin{table}[t]
	\centering
	\small
	\begin{tabular}{lll}
		\toprule
		Method & 7B & 13B \\
		\midrule
		FP16 & 5.46 & 4.89 \\
		AWQ (MLSys 24) & 5.65 & 5.00 \\
		DAQ & \underline{5.60} & \underline{4.97} \\
		AWQ+DAQ & \textbf{5.58} & \textbf{4.96} \\
		\midrule
		Improvement & 14.29\% & 12.50\% \\
		\bottomrule
	\end{tabular}
	\caption{Perplexity comparison of DAQ integration with AWQ on LLaMA-2. Lower perplexity indicates better performance. The best results are \textbf{bolded}. The second-best results are \underline{underlined}. }
	\label{tab:integration_results}
\end{table}

\textcolor{black}{As shown in Table \ref{tab:integration_results}}, integrating DAQ with AWQ consistently outperforms either method alone across different model sizes. This synergy stems from \textcolor{black}{ the complementary nature of these two methods:} AWQ uses per-channel scaling to protect salient weights, while DAQ leverages the non-uniform properties of FP representation to adjust the dynamic range, further preserving \textcolor{black}{salient} weights in FP quantization. 

% The superior performance of the combined AWQ+DAQ method demonstrates that DAQ can serve as an effective complement to existing quantization methods in FP quantization, offering a valuable enhancement to current weight-only quantization methods.

The superior performance of the combined AWQ+DAQ method demonstrates that DAQ provides a novel perspective on FP quantization, serving as an effective complement to existing weight-only quantization methods. By leveraging the non-uniform properties of FP representation and considering the density and impact of weights, DAQ can offer a valuable enhancement to existing weight-only quantization methods.

% \textcolor{black}{Notably, DAQ employs the same inference process as vanilla RTN, focusing solely on optimizing quantization parameters. This design enables DAQ to seamlessly integrate with and enhance the performance of advanced quantization methods.} 

% \subsection{}

% As quantization may adversely affect LLMs' performance on challenging downstream tasks, particularly in coding and mathematical domains, we evaluated DAQ on specialized models: WizardCoder-7B for coding tasks and MetaMath-7B for mathematical reasoning. Table X presents the results on the HumanEval benchmark for WizardCoder-7B and the GSM8K benchmark for MetaMath-7B.

% \begin{table}[htbp]
	% \centering
	% \caption{Comparison of DAQ with different variant models.}
	% % \begin{adjustbox}{width=.45\textwidth}
		% % \begin{tiny}
			% \begin{tabular}{lc}
				% \toprule
				% Method & Perplexity(↓) \\
				% \midrule
				% DAQ-MinMax & 6.65 \\
				% DAQ-Percentile & 6.89 \\
				% DAQ-DCA & 6.37  \\
				% DAQ-LDRA & 5.63  \\
				% DAQ w/o Zero-point & 5.65 \\
				% DAQ w/o Scale & 5.68 \\
				% DAQ & 5.60 \\
				% \bottomrule
				% \end{tabular}
			% % \end{tiny}
		% \label{ablation}
		% % \end{adjustbox}
	% \end{table}

% These findings underscore DAQ's effectiveness in preserving model performance on complex, domain-specific tasks. The significant improvements observed in both coding and mathematical reasoning tasks indicate that DAQ's density-aware approach to quantization is particularly beneficial for maintaining the nuanced capabilities of specialized LLMs.

\section{\textcolor{black}{Computational Complexity}}

The effectiveness of DAQ comes with certain computational considerations during the quantization process. The primary computational overhead stems from LDRA, which involves iterative optimization of quantization parameters. For each weight group, LDRA performs $T$ iterations of gradient descent, each requiring two forward passes through the layer to compute the finite difference. Given $N$ weight groups in a single layer, this results in a time complexity of $O(NTL)$, where $L$ represents the time for a single layer forward pass.
In practice, we observed that convergence is typically achieved with a relatively small number of iterations ($T \leq 1000$), keeping the overall computational cost manageable even for large-scale models. Moreover, the design of DAQ allows for parallel quantization across multiple layers, which can significantly reduce the total quantization time.

It is important to note that the additional computation required by DAQ is confined to the offline quantization process. During model inference, DAQ preserves the runtime performance of the quantized model, introducing no additional computational or storage overhead compared to vanilla RTN. This characteristic makes DAQ particularly suitable for resource-constrained \textcolor{black}{environments}, where improved quantization performance is desired without compromising inference efficiency. 

% In addition, this design allows for the incorporation of DAQ into other weight-only quantization methods, \textcolor{black}{further} enhancing their performance.

\section{\textcolor{black}{Conclusions}}
In this paper, we propose DAQ, a density-aware post-training weight-only quantization method. To leverage the non-uniform properties of floating-point representation, DAQ takes both the density and impact of weights into consideration. Specifically, DCA is introduced to align high-density weight regions with FP high-precision regions. Then, LDRA is employed to further adjust the dynamic range by optimizing the quantization parameters based on the impact of weights on the model output. Comprehensive experiments are conducted on LLaMA and LLaMA-2, and the results demonstrate the superiority of DAQ over state-of-the-art methods across various model sizes, quantization granularities, and calibration dataset sizes.

% 
% In the future, we will extend this work in the following directions. Firstly, we will enhance DCA by exploring more advanced density estimation methods (e.g., kernel density estimation) to \textcolor{black}{improve the identification of high-density weight regions.} Secondly, we will improve LDRA by introducing efficient gradient estimation methods (e.g., simultaneous perturbation stochastic approximation) to reduce computational overhead or improve gradient estimation accuracy. 

\newpage

\bibliography{aaai25}

\clearpage
\end{document}